\documentclass[10pt,twocolumn,letterpaper]{article}

\usepackage[pagenumbers]{cvpr} 

\usepackage{graphicx}
\usepackage{amsmath}
\usepackage{amssymb}
\usepackage{booktabs}
\usepackage{color, colortbl}
\definecolor{LightCyan}{rgb}{0.88,1,1}

\usepackage{hyperref}  
\usepackage[capitalize]{cleveref}
\crefname{section}{Sec.}{Secs.}
\Crefname{section}{Section}{Sections}
\Crefname{table}{Table}{Tables}
\crefname{table}{Tab.}{Tabs.}

\usepackage{multirow}
\usepackage{pifont}
\newcommand{\cmark}{\ding{51}}%
\newcommand{\xmark}{\ding{55}}%

\def\cvprPaperID{9784} 
\def\confName{CVPR}
\def\confYear{2023}

\begin{document}

\title{Self-Supervised Image-to-Point Distillation via Semantically Tolerant Contrastive Loss}
\author{
Anas Mahmoud$^1$, Jordan S. K. Hu$^1$, Tianshu Kuai$^1$, Ali Harakeh$^2$,  Liam Paull$^2$, and Steven L. Waslander$^1$\\
$^1$University of Toronto Robotics Institute, $^2$Mila, Université de Montréal\\
}
\maketitle

\begin{abstract}
   An effective framework for learning 3D representations for perception tasks is distilling rich self-supervised image features via contrastive learning.
    However, image-to-point representation learning for autonomous driving datasets faces two main challenges: 1) the abundance of self-similarity, which results in the contrastive losses pushing away semantically similar point and image regions and thus disturbing the local semantic structure of the learned representations, and 2) severe class imbalance as pretraining gets dominated by over-represented classes.  
   We propose to alleviate the self-similarity problem through a novel semantically tolerant image-to-point contrastive loss that takes into consideration the semantic distance between positive and negative image regions to minimize contrasting semantically similar point and image regions. Additionally, we address class imbalance by designing a class-agnostic balanced loss that approximates the degree of class imbalance through an aggregate sample-to-samples semantic similarity measure.
   We demonstrate that our semantically-tolerant contrastive loss with class balancing improves state-of-the-art 2D-to-3D representation learning in all evaluation settings on 3D semantic segmentation. Our method consistently outperforms state-of-the-art 2D-to-3D representation learning frameworks across a wide range of 2D self-supervised pretrained models.
\end{abstract}

\section{Introduction}
\label{sec:intro}
\begin{figure}[t]
  \centering
  \includegraphics[scale=0.42]{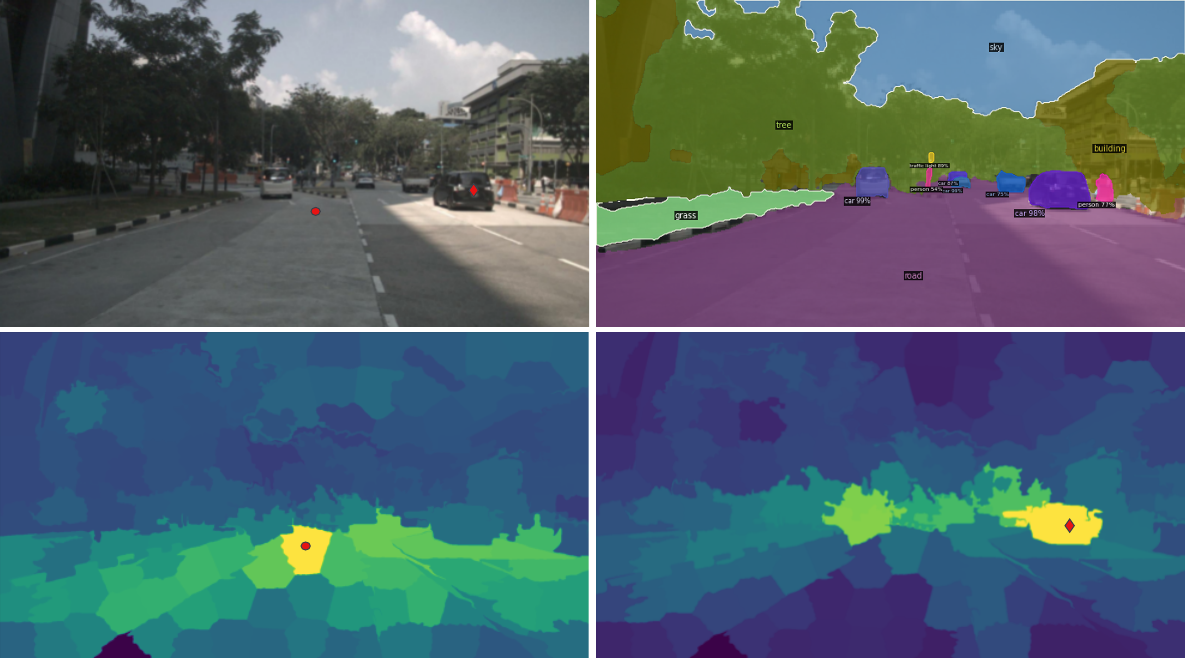}
  \caption{\textbf{Bottom row:} Superpixel-to-superpixel cosine similarity with respect to, \textbf{bottom left:} a road anchor, and \textbf{bottom right:} a vehicle anchor (both marked in red). Superpixel-driven contrastive loss~\cite{sautier2022slidr} treats all superpixels excluding the anchor as negative samples. As such, loss will be dominated by gradients from semantically similar negative samples, disturbing the local semantic structure of the learned 3D representations. Our loss uses superpixel similarity to 1) Reduce the contribution of false negative samples, and 2) Balance the contribution of well-represented (i.e., road) and under-represented (i.e., vehicle) anchors}
  \label{fig:slidr_contrastive_loss}
\end{figure}
Self-supervised learning (SSL) has shown significant success in learning useful representations from unlabeled images~\cite{chen2020_simclr, caron2020_swav, sun2020waymo, gidaris2021obow}, mainly due to large, diverse, and balanced 2D image datasets. These successes promise to alleviate the requirement for large labeled datasets, which can be expensive, not attainable, or task-specific. These issues are exacerbated when generating labels for 3D point clouds, which are usually much more difficult to annotate~\cite{xie2020pointcontrast} than 2D images. Additionally, the sparse nature of point clouds generated using a LiDAR sensor, as is common in outdoor autonomous driving data, substantially increases the difficulty of manually generating per-point labels, particularly at large distances.

A common approach to learn 3D representations is through multimodal invariance~\cite{li2022closer}, where 3D representations are learned to be invariant to  features extracted from image encoders trained with self-supervised learning~\cite{liu2021ppkt, sautier2022slidr}. The current state-of-the-art, SLidR~\cite{sautier2022slidr}, encourages learning representations of 3D point regions (superpoints) to match pre-trained representations of 2D image regions (superpixels) through a novel contrastive loss. By contrasting 2D and 3D regions, SLidR~\cite{sautier2022slidr} enables learning representations from point clouds with varying point densities, as is common in autonomous driving datasets.

Unfortunately, SlidR's region-based sampling does not address self-similarity, which results when fewer unique semantic classes exist in the data relative to the number of chosen contrastive pairs during training. Under self-similarity, many negative samples will belong to the same semantic class as the positive sample used to compute the contrastive loss, pushing apart their embeddings and breaking the local semantic structure of learned 3D representation~\cite{wang2021understanding}(see~\Cref{fig:slidr_contrastive_loss}). This issue is further exacerbated by the implicit hardness-aware property of contrastive loss, where the largest gradient contributions come from the most semantically similar negative samples~\cite{wang2021understanding}(see~\ref{subsec:limiations}). 

In addition, autonomous driving datasets are highly imbalanced, for example, in the nuScenes dataset~\cite{caesar2020nuscenes}, the 'Pedestrian' class covers 0.25\% of the data, while  'vegetation' class covers 22.19\% of the data. Since the class of the positive sample is unknown during pretraining, SLidR's loss gives an equal weight to all samples in the batch. Hence, the 3D pretraining is predominately driven by gradients from a few over-represented samples, leading to poor performance on under-represented samples. 

\begin{figure}[t!]
  \centering
  \includegraphics[scale=0.55]{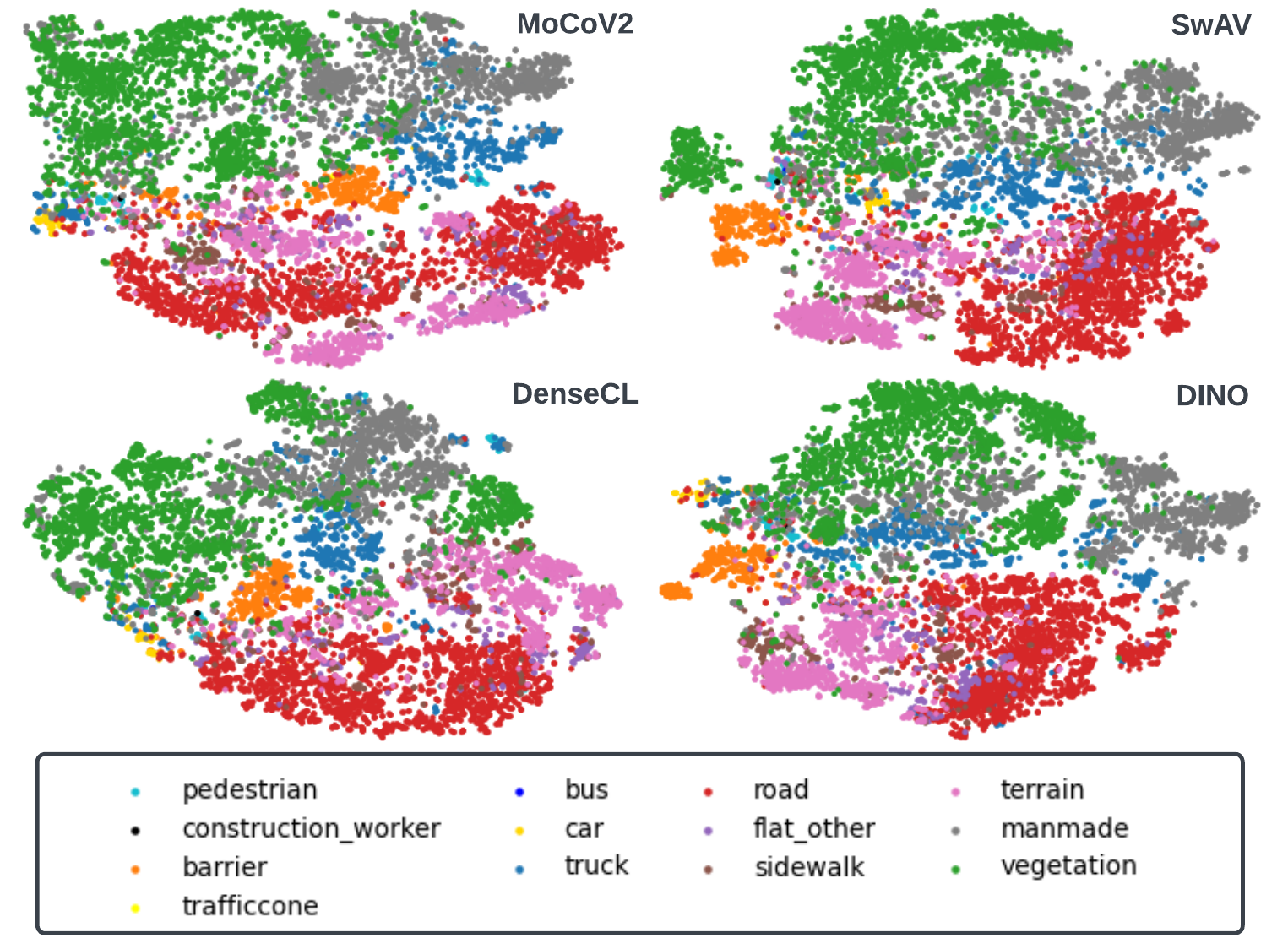}
  \caption{t-SNE~\cite{tsne} visualization of superpixel-level features for a given batch of nuScenes~\cite{caesar2020nuscenes} images. Each superpixel feature is colorized based on its semantic class derived from LiDAR ground truth point-wise labels. Here, we show that MoCoV2~\cite{chen2020_mocov2}, SwAV~\cite{caron2020_swav}, DenseCL~\cite{wang2021_densecl} and DINO~\cite{caron2021emerging_dino} weights generate meaningful semantic clusters on the superpixel level.}
  \label{fig:superpixel_features}
\end{figure}

In this work, we simultaneously address the challenge of contrasting semantically similar point and image regions and the challenge of learning 3D representations from highly imbalanced autonomous driving datasets. \Cref{fig:slidr_contrastive_loss} shows that image regions semantically similar to the anchor exhibit high cosine similarity in the 2D self-supervised feature space. Our first key idea is to exploit the semantic distance between positive and negative pooled image features to guide negative sample selection. Reducing the contribution of false negative samples, which are abundant in autonomous driving datasets due to the self-similarity, prevents the disturbance of the local semantic structure of the pre-trained 3D representations~\cite{wang2021understanding}. \Cref{fig:superpixel_features} shows that most anchors come from over-represented classes (i.e., road, vegetation) resulting in a 3D point encoder that is less discriminative with respect to under-represented classes. To address this challenge, we propose using aggregate semantic similarity between samples as a proxy for class imbalance. By balancing the contribution of over and under-representated anchors, we improve the learned 3D representations of under-represented semantic classes (i.e., pedestrians and vehicles). We summarize our approach with two main contributions, which we present below.

\noindent\textbf{Semantically-Tolerant Loss}. To address the similarity of samples in 2D-to-3D representation learning frameworks, we propose a novel contrastive loss that relies on 2D self-supervised image features to infer the semantic distance between positive and negative pooled image features. We propose to either directly reduce the gradient contribution of semantically-similar negative samples or exclude the K-nearest samples based on the semantic distance to the positive sample. 

\noindent \textbf{Class Agnostic Balanced Loss}. To address  pre-training using highly imbalanced 3D data, we propose a novel class agnostic balancing for contrastive losses that weights the contribution of each 3D region in a point cloud based on the aggregate semantic similarity of its corresponding 2D region with all negative samples. We reason that samples with high aggregate semantic similarity to other samples come from over-represented classes, while under-represented samples are semantically similar to very few other samples. Hence, we reduce the contribution of over-represented samples, while increasing that of under-represented samples. 

By extending the  state-of-the-art 2D-to-3D representation learning frameworks using our proposed semantically-tolerant contrastive loss with class balancing, we show that we can improve their performance on in-distribution linear probing and finetuning semantic segmentation, as well as on out-of-distribution few-shot semantic segmentation. We also show that our proposed semantically-tolerant loss improves 3D semantic segmentation performance across a wide range of 2D self-supervised pretrained image features, consistently outperforming state-of-the-art 2D-to-3D representation learning frameworks.

\section{Related Work}
\label{sec:related_work}
\subsection{Self-Supervised 2D Representation Learning}
Learning 2D representations via instance-level discrimination has shown to be an effective pre-text task for SSL frameworks~\cite{wu2018unsupervised_instance}. These frameworks learn unsupervised representations by maximizing the mutual information~\cite{bachman2019learning} between two augmented views of an image using one of two objectives. Similarity maximization objective including contrastive methods~\cite{chen2020_simclr, chen2020_mocov2, wang2021_densecl} minimize the distance between the representations of the two views of the same instance, while maximizing the distance to other instances. On the other hand, redundancy minimization~\cite{zbontar2021barlow, bardes2022vicreg} objective minimizes the statistical correlation between the dimensions of the learned representation while also maximizing similarity between representations of the same instance. SSL frameworks can also be categorized based on whether their losses are designed to discriminate pixel-level~\cite{wang2021_densecl}, superpixel-level~\cite{henaff2021efficient_detcon} or instance-level~\cite{chen2020_simclr} representations. DetCon~\cite{henaff2021efficient_detcon} has demonstrated that designing contrastive losses on image regions named superpixels leads to efficient pretraining and better performance especially on dense tasks like 2D semantic segmentation and object detection compared to instance-level contrastive losses. 
 
In this paper, we employ the self-supervised 2D representations as a supervisory signal to learn 3D representations. We demonstrate that the semantic structure learned by 2D SSL methods can be used to address challenges in learning 3D representations for autonomous driving datasets including abundance of self-similarity and severe class imbalance.

\subsection{Self-Supervised 3D Representation Learning}
Self-supervised 3D representation learning can be categorized into perspective-invariant, format-invariant and multimodal invariant methods~\cite{li2022closer}. Perspective invariant methods like PointContrast~\cite{xie2020pointcontrast} learn point representations that are invariant to different views of the point cloud. These methods are primarily designed for indoor RGB-D datasets, where full 3D reconstruction of the scene is possible~\cite{Zhang_2021_depthcontrast}. To address the limitation of requiring multiple views, DepthContrast~\cite{Zhang_2021_depthcontrast}, a format-invariant method, uses only single view point cloud data, and learns 3D representations that are invariant to point and voxel representations. By design, the contrastive loss in DepthContrast~\cite{Zhang_2021_depthcontrast} learns global scene-level representations and therefore is prone to losing information on small objects~\cite{sautier2022slidr}. 

Multi-modal invariant methods extract image representations from pretrained image encoders, and use a contrastive loss to learn 3D representations by maximizing the similarity to 2D representations. PPKT~\cite{liu2021ppkt} contrasts representations of pairs of point and pixel correspondences. This method is mainly designed for indoor RGB-D datasets, where dense point-to-pixel correspondences exist~\cite{sautier2022slidr}. SLidR~\cite{sautier2022slidr} is the first 3D representation learning method that is primarily designed for autonomous driving datasets. Inspired by DetCon~\cite{henaff2021efficient_detcon}, they use unsupervised image segmentation algorithms~\cite{slic} to segment images into superpixels. By projecting the point cloud onto the superpixel mask, each point is assigned a superpoint. Each superpoint and its corresponding superpixel form a positive pair and the contrastive loss is thus defined at the superpixel-level. SLidR~\cite{sautier2022slidr} formulation has multiple advantages; First, constructing semantically coherent image and point cloud corresponding regions, leads to learning useful object-level representations~\cite{henaff2021efficient_detcon}. In addition, unlike PPKT~\cite{liu2021ppkt}, grouping superpoint and superpixel representations using average pooling increases robustness against camera and LiDAR calibration errors~\cite{sautier2022slidr}. Finally, the density of LiDAR returns increases as a function of distance~\cite{Hu_2022_PDV, DVF, bounding_box_painting} resulting in very few number of points at mid-to-long range objects. The Random sampling strategy of positive pairs in PointContrast~\cite{xie2020pointcontrast} and PPKT~\cite{liu2021ppkt} applied to outdoor scenes results in a biased sampling towards dense nearby points. SLidR~\cite{sautier2022slidr} breaks down the scene based on image superpixels, which leads to constrastive pairs covering the entire 3D scene. Each pair has the same weight in the contrastive loss regardless of the number of points in these regions~\cite{sautier2022slidr}. 

The contrastive losses proposed in PPKT~\cite{liu2021ppkt} and SLidR~\cite{sautier2022slidr} are not suited for autonoumous driving data due to two main reasons. First, the high level of self-similarity, which results in the number contrastive pairs from unique semantic classes being much less than the number of pairs in any given batch. This phenomenon will lead to the contastive loss considering semantically similar samples as negative samples and pushing their representations apart. Second, autonomous driving datasets suffer from severe class imbalance which can lead to a small over-represented subset of the semantic class dominating the self-supervised pretraining stage. We address the first challenge by formulating a semantically tolerant loss which prevents contrasting semantically similar samples. We address the second challenge by formulating a class-agnostic balanced loss that uses aggregate sample-to-samples semantic distance as a proxy for class imbalance. 

\section{Methodology}
\label{sec:methodology}
\subsection{Superpixel-Driven Contrastive Loss}

\begin{figure*}[t!]
  \centering
  \includegraphics[width=0.8\textwidth]{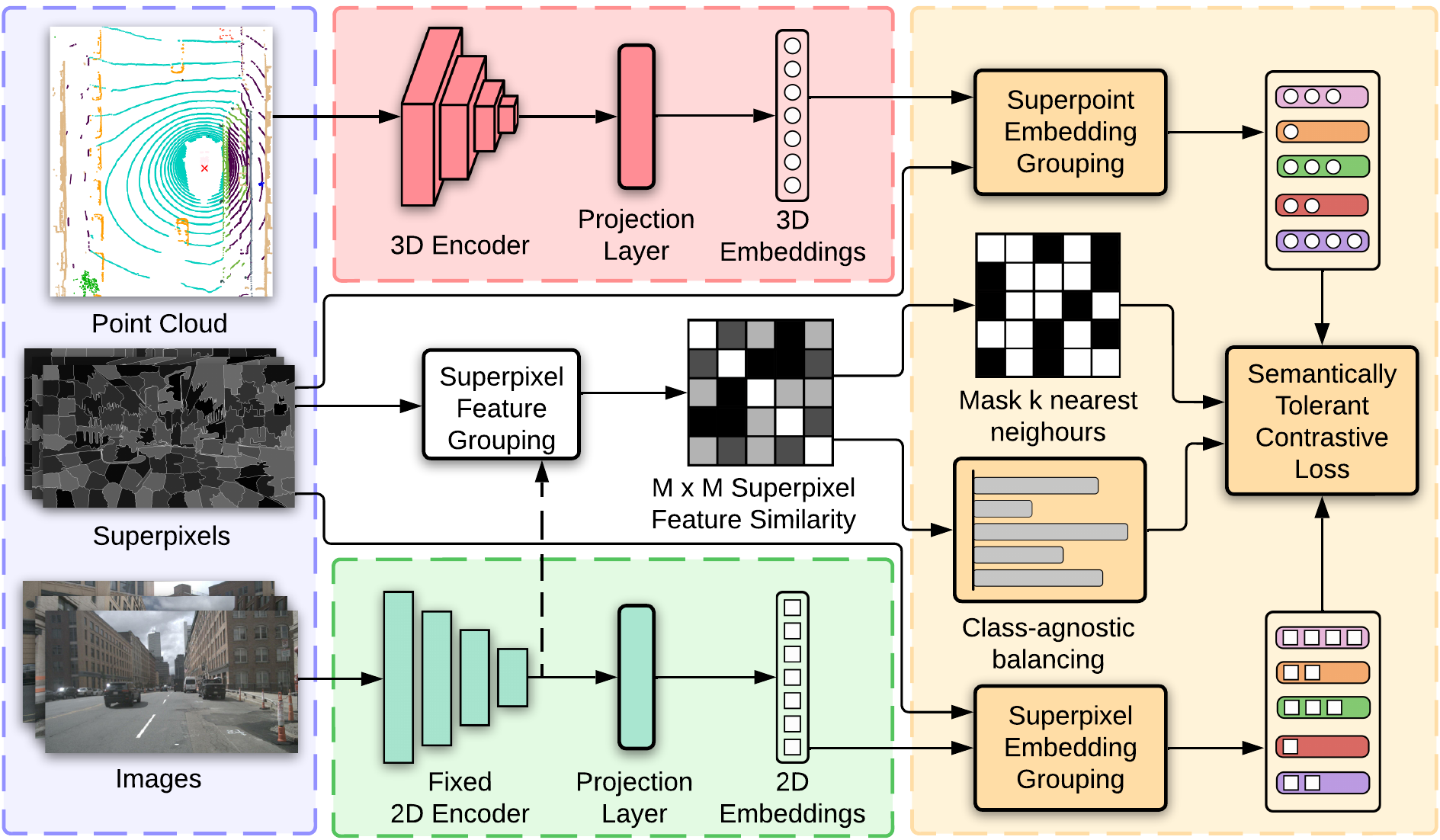}

  \caption{An overview of our self-supervised image-to-point distillation framework. LiDAR and camera data are encoded and their respective features are projected onto an embedding space where the contrastive loss is computed. Then, superpixels are used to group the 3D and 2D embeddings. In addition, the fixed 2D self-supervised features are grouped and used to estimate the superpixel-to-superpixel similarity. Finally, our loss utilizes the  similarity estimates to: 1) Reduce the contribution of false negative superpixel embeddings by masking the k-nearest neigbouring superpixels, and 2) Balance the contribution of over and under-represented samples based on the distribution of the aggregate superpixel feature similarities.}
  \label{fig:sem-slidr-overview}
\end{figure*}

\subsubsection{Background}
Given a set of point clouds representing multiple scenes $\left\{\mathbf{p}_i=\left\{\boldsymbol\ell_i, \mathbf{f}_i\right\} \mid i=1, \ldots, U\right\}$, where $\boldsymbol\ell_i \in \mathbb{R}^{N_i \times 3}$ is a tensor of 3D location of $N_i$ points representing the $i^{th}$ scene, and $\mathbf{f}_i \in \mathbb{R}^{N_i \times L}$ are their associated point-wise features (i.e., intensity and elongation). We also have a set of camera-to-LiDAR synchronized images $\left\{ \left\{ \mathbf{I}_i^1, \ldots, \mathbf{I}_i^J\right\} \mid i=1, \ldots, U \right\}$, where $\mathbf{I}_i^j \in \mathbb{R}^{H \times W \times 3}$ and $J$ is the number of cameras per scene. Using the unsupervised segmentation algorithm SLIC \cite{slic}, each pixel of image $\mathbf{I}_i^j$ is segmented into a set of superpixels $\mathcal{X}_i^j$ and $\left\{ \mathcal{X}_i = \left\{ \mathcal{X}_i^1, \ldots, \mathcal{X}_i^J\right\} \mid i=1, \ldots, U \right\}$ denotes the set of multi-scene superpixels $\mathcal{X}$. To generate the set of corresponding superpoints $\left\{ \mathcal{P}_i = \left\{ \mathcal{P}_i^1, \ldots, \mathcal{P}_i^J\right\} \mid i=1, \ldots, U \right\}$, camera-to-LiDAR calibration matrices are used to map 3D points to pixel locations. Superpixels with no corresponding superpoints (i.e., superpixels outside FOV of the LiDAR) are removed from set $\mathcal{X}$ and therefore $|\mathcal{X}| = |\mathcal{P}|$. 

Let the point cloud encoder be a 3D deep neural network $f_{\theta_{P}} : \mathbb{R}^{N \times (3 + L)} \to \mathbb{R}^{N \times D}$, with randomly initialized, trainable parameters $\theta_{P}$. Let the image encoder be a 2D neural network $g_{\theta_{I}} : \mathbb{R}^{H \times W \times 3} \to \mathbb{R}^{\frac{H}{s} \times \frac{W}{s} \times C}$, with parameters $\theta_{I}$ initialized from any 2D self-supervised pre-trained parameters. The goal is to use the pretrained image features at the superpixel level as a supervisory training signal for the point cloud encoder. 

To compute the superpixel-driven contrastive loss, a trainable projection layer $h_{\omega_P}: \mathbb{R}^{N \times D} \to \mathbb{R}^{N \times E}$ maps the output of the point cloud encoder to the contrastive loss embedding space. In addition, a trainable projection layer $h_{\omega_I}: \mathbb{R}^{\frac{H}{s} \times \frac{W}{s} \times C} \to \mathbb{R}^{\frac{H}{s} \times \frac{W}{s} \times E}$ maps pixel-level image features to the contrastive loss embedding space. The loss is computed between superpoint and superpixel embeddings. First, $\mathcal{P}$ and $\mathcal{X}$ are used to group point and pixel embeddings respectively. Let $M = |\mathcal{X}| = |\mathcal{P}|$. An average pooling function is then applied to the point and pixel embeddings within each group, to extract multi-scene superpoint embeddings $\mathbf{Q} \in \mathbb{R}^{M \times E}$ and superpixel embeddings $\mathbf{K} \in \mathbb{R}^{M \times E}$. The superpixel-driven loss is formulated as:
\begin{align}
    \mathcal{L\left(\mathbf{Q}, \mathbf{K}\right)}=- \frac{1}{M} {\sum_{i=0}^{M} \log \left[\frac{e^{\left(\left<\mathbf{q}_i,\mathbf{k}_i\right>/\tau\right)}}{\sum_{j \neq i} e^{\left(\left<\mathbf{q}_i,\mathbf{k}_j\right>/\tau\right)}+e^{\left(\left<\mathbf{q}_i,\mathbf{k}_i\right>/\tau\right)}}\right]},
    \label{eq:loss_slidr}
\end{align}
where $\left<\mathbf{q}_i,\mathbf{k}_j\right>$ is a measure of similarity computed as the dot product between the $\ell_2$-normalized superpoint and superpixel embeddings and $\tau$ is the temperature scale~\cite{chen2020_simclr}. 

\subsubsection{Limitations}
\label{subsec:limiations}
\noindent\textbf{Self-similarity in Autonomous Driving Data}. Looking at~\Cref{fig:slidr_contrastive_loss}, we observe that many image regions defined by superpixels belong to the same semantic class. We call this self-similarity. For a given batch, each superpoint and its corresponding superpixel embedding are considered positive samples, while the remaining pairs are treated as negative samples. Due to the self-similarity, there is a high probability for the contrastive loss in SLidR~\cite{sautier2022slidr} (and a higher probability in PPKT~\cite{liu2021ppkt}) of pushing away semantically similar samples.

\vspace{0.05in}
\noindent\textbf{Hardness-aware Property of Contrastive Loss}. The success of the softmax-based contrastive loss has been attributed to its hardness-aware property~\cite{wang2021understanding}. The temperature parameter $\tau$ controls the distribution of negative gradients, where low temperatures lead to larger gradient contribution from nearest neighbour negative samples. Authors in~\cite{wang2021understanding} have demonstrated that there exists a uniformity-tolerance dilemma in softmax-based contrastive losses. They show that high temperatures lead to semantically tolerant embeddings, but can suffer from embedding collapse, while low temperatures lead to a uniform distribution of embeddings, preventing embedding collapse. Nonetheless, low temperatures also lead to less tolerant embeddings, where similar samples are not closely clustered. Both PPKT~\cite{liu2021ppkt} and SLidR~\cite{sautier2022slidr} use a low temperature of $\tau=0.04$ and $\tau=0.07$ respectively. Due to the implicit hardness-aware property of the contrastive loss, the highest gradient contribution to the pre-training signal comes from pushing away semantically similar samples which disturbs the local semantic structure of the learned representation.

\subsection{Semantically-Tolerant Contrastive Loss}
We observe that in self-supervised image-to-point cloud knowledge distillation frameworks~\cite{liu2021ppkt, sautier2022slidr}, pre-trained models provide a strong prior on the relationship between superpixel features.  To show this, we use ResNet-50 pre-trained on ImageNet~\cite{russakovsky2015imagenet} using four 2D self-supervised methods. First, we use each pre-trained model to map a batch of 16 images from nuScenes dataset~\cite{caesar2020nuscenes} to output features $g_{\theta_{I}} : \mathbb{R}^{H \times W \times 3} \to \mathbb{R}^{\frac{H}{s} \times \frac{W}{s} \times C}$. Using $\mathcal{X}$, pixel features are then grouped and an average pooling function is applied to extract superpixel features $\mathbf{F} \in  \mathbb{R}^{M \times C}$. To visualize $\mathbf{F}$, we reduce the dimensionality of the features using t-SNE~\cite{tsne} and show the first 2 dimensions in~\Cref{fig:superpixel_features}. Each point corresponds to a superpixel feature $\mathbf{f}_i \in {M \times C}$ colorized using the dominant point-wise ground truth label of their corresponding superpoint. 

As seen in ~\Cref{fig:superpixel_features}, extracted superpixel features from nuScenes~\cite{caesar2020nuscenes} images form relatively coherent semantic clusters. SLidR~\cite{sautier2022slidr} not only ignores this strong prior in their contrastive loss formulation but also suffers due to the hardness-aware property~\cite{wang2021understanding} of the contrastive loss, which will in its current form, primarily focus on pushing away semantically similar superpoints and superpixels embeddings. In addition, we observe a high level of self-similarity, where the ratio between the total number of superpixels (i.e., 9000) and the number of unique semantic classes (i.e., 13) in a batch is very high, leading to an increase in false negatives.

\vspace{0.05in}
\noindent\textbf{Similarity-aware Loss}.
Our goal is to address the issue of contrasting semantically similar superpoint and superpixel embeddings. This issue is exacerbated due to the high self-similarity in autonomous driving data, and the hardness-aware property of the contrastive loss. We propose a semantically-tolerant contrastive loss that utilizes superpixel similarities in the feature space to re-weight the contribution of semantically similar negative samples. Our loss reduces the gradient contribution from negative superpixel embeddings that are semantically similar to the positive sample.  Our loss can be written as:
\begin{multline}
    \mathcal{L}_{\alpha}\left(\mathbf{Q}, \mathbf{K}\right)= \\
    - \frac{1}{M} {\sum_{i=0}^{M} \log \left[\frac{e^{\left(\left<\mathbf{q}_i,\mathbf{k}_i\right>/\tau\right)}}{\sum_{j \neq i} e^{\left((1 - \alpha_{ij}) \ . \ \left<\mathbf{q}_i,\mathbf{k}_j\right>/\tau\right)}+e^{\left(\left<\mathbf{q}_i,\mathbf{k}_i\right>/\tau\right)}}\right]},
    \label{eq:loss_alpha}
\end{multline}
where $\alpha_{ij} = \left<\mathbf{f}_i, \mathbf{f}_j\right>$ is a measure of similarity defined as the dot product of the $\ell_2$ normalized superpixel features $\mathbf{F}$. Here, $\mathbf{f}_i \in \mathbb{R}^{C}$. 
\begin{align}
    P_{ij}=\left[\frac{e^{\left((1 - \alpha_{ij}) \ . \ \left<\mathbf{q}_i,\mathbf{k}_j\right>/\tau\right)}}{\sum_{j \neq i} e^{\left((1 - \alpha_{ij}) \ . \ \left<\mathbf{q}_i,\mathbf{k}_j\right>/\tau\right)}+e^{\left(\left<\mathbf{q}_i,\mathbf{k}_i\right>/\tau\right)}}\right].
\end{align}
Looking at the gradient of the proposed loss with respect to the negative similarity $\left<\mathbf{q}_i,\mathbf{k}_j\right>$ where $j \neq i$:
\begin{align}
\frac{\partial \mathcal{L}_{\alpha}\left(\mathbf{q}_i, \mathbf{K}\right)}{\partial \left<\mathbf{q}_i,\mathbf{k}_j\right>} = \frac{(1-\alpha_{ij})}{\tau} P_{ij}.
\end{align}
In $\mathcal{L}_{\alpha}$, the gradient of the loss with respect to the negative sample $\mathbf{k}_j$, is weighted by the dissimilarity between the features of positive superpixel $\mathbf{f}_i$ and the negative superpixel $\mathbf{f}_j$. Therefore, there is less contribution from $\mathbf{k}_j$'s that are semantically similar to $\mathbf{k}_i$. When $\alpha_{ij}= 1.0$ the pos/neg superpixel features are identical, and $\frac{\partial \mathcal{L}_{\alpha}\left(\mathbf{q}_i, \mathbf{K}\right)}{\partial \left<\mathbf{q}_i,\mathbf{k}_j\right>} = 0.0$ preventing contrasting semantically similar pairs. When $\alpha_{ij}= 0.0$ then $\frac{\partial \mathcal{L}_{\alpha}\left(\mathbf{q}_i, \mathbf{K}\right)}{\partial \left<\mathbf{q}_i,\mathbf{k}_j\right>} = \frac{P_{ij}}{\tau}$ and we revert back to the SLidR~\cite{sautier2022slidr} formulation.

\vspace{0.05in}
\noindent\textbf{Nearest-Neighbour-aware Loss}. Training with $\mathcal{L}_{\alpha}$ is a much easier loss to minimize compared to $\mathcal{L}$ since the closest negative samples which are also the hardest negatives have lower contribution to the loss. For instance, when the mean of superpixel-to-superpixel feature similarities is high for a given 2D SSL pretrained model, $\sum_{j \neq i} e^{\left((1-\alpha_{ij}) \ . \ \left<\mathbf{q}_i,\mathbf{k}_i\right>/\tau\right)} << e^{\left(\left<\mathbf{q}_i,\mathbf{k}_i\right>/\tau\right)}$ and thus very few negative samples will contribute to the loss. In this case, $\mathcal{L}_{\alpha}$ can be easily minimized without learning useful representations. One approach to address this issue, is to  suppress negative samples with low similarity values by setting $\alpha_{ij} < \alpha_{min}$ to 0.0, where $\alpha_{min}$ is a tunable parameter. Hence, we ensure we have enough negative samples to learn useful representations while preventing contrasting against semantically similar negative samples. However,  $\alpha_{min}$ has to be tuned for each 2D pretrained model. Empirically, we find that the linear separability of the 3D representations is very sensitive to the choice of $\alpha_{min}$ (see~\Cref{tab:cosine_sim_K_nearest}).  

We hypothesize that the values of superpixel-to-superpixel similarities are not well-calibrated and vary based on the 2D pretrained model. Therefore, these similarities should not be directly used in the contrastive loss. For a given positive sample, we hypothesize that the order of similarities across different 2D pretrained models is consistent. Hence, the order of similarities is more informative than the similarity values. To avoid directly incorporating un-calibrated values of $\alpha_{ij}$ in the loss, we propose removing the $K$-nearest neighbours from the set of negative samples based on $\alpha_{ij}$. To this end, for each postive sample $\mathbf{q}_i$, we sort $\alpha_{ij}, \ \forall j\neq i$ and compute $\alpha_{iK}$ that contains the $K$-nearest neighbours. Here, $C_{ij}$ is an indicator of whether $\mathbf{f}_j$ is semantically similar to $\mathbf{f}_i$.
\begin{equation}
  C_{ij}=\left\{
  \begin{array}{@{}ll@{}}
    1.0, & \text{if}\ \alpha_{ij} < \alpha_{iK} \\
    0.0, & \text{otherwise}
  \end{array}\right.
\end{equation} 
The loss can then be formulated as:
\begin{multline}
        \mathcal{L}_{knn}\left(\mathbf{Q}, \mathbf{K}\right)= \\ 
        - \frac{1}{M} {\sum_{i=0}^{M} \log \left[\frac{e^{\left(\left<\mathbf{q}_i,\mathbf{k}_i\right>/\tau\right)}}{\sum_{j \neq i} C_{ij} \ . \ e^{\left(\left<\mathbf{q}_i,\mathbf{k}_j\right>/\tau\right)}+e^{\left(\left<\mathbf{q}_i,\mathbf{k}_i\right>/\tau\right)}}\right]}.
    \label{eq:loss_nearest_neigbour}
\end{multline}
Using $\mathcal{L}_{knn}$, a fixed number of negative samples, excluding the $K$-nearest neighbors are used as the negative set. We show in~\Cref{tab:cosine_sim_K_nearest} that excluding the $K$-nearest neighbors results in better linear separability compared to directly incorporating $\alpha_{ij}$ in the contrastive loss.

\subsection{Class-Agnostic Balanced Contrastive Loss}
Autonomous driving datasets are highly imbalanced (see~\Cref{fig:superpixel_features}), for instance, in nuScenes~\cite{caesar2020nuscenes} only $0.05\%$ of the superpixels belong to 'motorcycle' and 'bicycle' categories, while $45\%$ of the superpixels belong to 'road' and 'vegetation' categories. For indoor 3D point cloud datasets, the problem is less severe, where the rarest category 'sink' in ScanNet V2~\cite{chen2020_mocov2} consists of $2.75\%$ of the points~\cite{jiang2021guided}. Since the category of a sample is unknown during pre-training, PPKT~\cite{liu2021ppkt} and SLidR~\cite{sautier2022slidr} assume a fixed weight of $\frac{1}{M}$ on the loss from each sample within a batch. Since the training signal is dominated by gradients from samples of over-represented categories, the learned representations for under-represented categories might not be optimal. 

To address this issue, we reason that superpixel-to-superpixel similarity can also be used as a proxy for class imbalance. For example, for an over-represented anchor $\mathbf{q}_i$ in a batch, its associated superpixel feature $\mathbf{f}_i$ will have high $\alpha_{ij}$ with a large number of negative samples, while an under-represented anchor, will have low $\alpha_{ij}$ with most negative samples. To balance the training, first, we compute votes for each anchor based on similarity $v_i = \sum_{j=1}^{M} \alpha_{ij}$. Then, a min-max normalization is applied $v_i = \frac{v_i - v_{min}}{v_{max}}$ to suppress noise. Finally, for each anchor $\mathbf{q}_i$, we assign a weight $w_i$ inversely proportional to $v_i$. The semantically-tolerant loss with class-agnostic balancing can be formulated as:
\begin{multline}
        \mathcal{L}_{ST}\left(\mathbf{Q}, \mathbf{K}\right)= \\ 
        - {\sum_{i=0}^{M} \frac{w_i}{w} \log \left[\frac{e^{\left(\left<\mathbf{q}_i,\mathbf{k}_i\right>/\tau\right)}}{\sum_{j \neq i} C_{ij} \ . \ e^{\left(\left<\mathbf{q}_i,\mathbf{k}_j\right>/\tau\right)}+e^{\left(\left<\mathbf{q}_i,\mathbf{k}_i\right>/\tau\right)}}\right]}.
    \label{eq:loss_nearest_neigbour}
\end{multline}
Where $w_i = 1.0 - v_i$ and $w = \sum_{i=0}^{M} w_i$. 
\section{Experiments} 
\label{sec:experiments}

\subsection{Pre-training}
\noindent\textbf{Backbones}.
To represent the input point cloud, we transform the 3D points from Cartesian coordinates $(x, y, z)$ to cylindrical coordinates $(\rho, \phi, z)$. The input is voxelized using 3D cylindrical partitioning~\cite{zhu2021cylindrical}, where the voxel size is $(\delta\rho=10cm, \delta\phi=1^{\circ}, \delta z=10cm)$. For the 3D backbone, similar to SLidR~\cite{sautier2022slidr}, we use the Minkowski U-Net with $3\times3\times3$ kernels for all sparse convolutional layers.
For the 2D backbone, we use the ResNet-50~\cite{he2016deep_resnet} architecture and initialize the model parameters using a multitude of 2D self-supervised pretrained models including MoCoV2~\cite{chen2020_mocov2}, SwAV~\cite{caron2020_swav}, DINO~\cite{caron2021emerging_dino}, OBoW~\cite{gidaris2021obow} and DenseCL~\cite{wang2021_densecl}. For all experiments except  2D SSL frameworks, the 2D backbone for PPKT~\cite{liu2021ppkt}, SLidR~\cite{sautier2022slidr} and ST-SLidR is initialized using MoCoV2~\cite{chen2020_mocov2}.

\vspace{0.05in}
\noindent\textbf{Dataset}.
For pre-training, we use the nuScenes~\cite{caesar2020nuscenes} dataset, which contains 700 training scenes.  Following SLidR~\cite{sautier2022slidr}, we further split the 700 scenes into 600 for pre-training and 100 scenes for selecting the optimal hyper-parameters. During pretraining, only keyframes from the 600 scenes are used to train both SLidR and ST-SLidR.

\vspace{0.05in}
\noindent\textbf{Training and Data Augmentations}. For all experiments, we pre-train the point cloud encoder, $f_{\theta_P}$, the superpoint embedding layer, $h_{\omega_P}$ and the superpixel embedding layer, $h_{\omega_I}$, for 50 epochs on 2 A100 GPUs with a batch size of 16. Similar to SLidR~\cite{sautier2022slidr}, we use an SGD optimizer with a momentum of 0.9,  an initial learning rate of 0.5 and a cosine annealing learning rate scheduler. Finally, for regularization, we use a weight decay of 0.0001 and dampening of 0.1. For data augmentation, we apply linear transformations to the point cloud including random flipping in the $x$ and $y$-axis, and rotating around $z$-axis. In addition, similar to DepthContrast~\cite{Zhang_2021_depthcontrast}, we randomly select a cube and drop all points within the cube. For images, we apply random crop-resize and horizontal flip. 
\begin{table}[t]
\centering
\resizebox{\columnwidth}{!}{
\begin{tabular}{lccccc}
\toprule
\multicolumn{1}{c}{\multirow{3}{*}{\begin{tabular}[c]{@{}c@{}}3D\\  Initialization\end{tabular}}} & \multicolumn{1}{c}{\multirow{3}{*}{Reference}} & \multicolumn{2}{c}{nuScenes}                                                                                                                                                                  & \multicolumn{1}{c}{KITTI}                                                                   & Waymo                                                                    \\ \cmidrule{3-6} 
\multicolumn{1}{c}{}                                                                              & \multicolumn{1}{c}{}                           & \multicolumn{1}{c}{\multirow{2}{*}{\begin{tabular}[c]{@{}c@{}}Lin. Prob\\ 100\%\end{tabular}}} & \multicolumn{1}{c}{\multirow{2}{*}{\begin{tabular}[c]{@{}c@{}}Finetune\\ 1\%\end{tabular}}} & \multicolumn{1}{c}{\multirow{2}{*}{\begin{tabular}[c]{@{}c@{}}Finetune\\ 1\%\end{tabular}}} & \multirow{2}{*}{\begin{tabular}[c]{@{}c@{}}Finetune\\  1\%\end{tabular}} \\
\multicolumn{1}{c}{}                                                                              & \multicolumn{1}{c}{}                           & \multicolumn{1}{c}{}                                                                           & \multicolumn{1}{c}{}                                                                        & \multicolumn{1}{c}{}                                                                        &                                                                          \\ \midrule
\multicolumn{1}{l}{Random}                                                                        & \multicolumn{1}{c}{N/A}                        & 8.10                                                                                            & 30.30                                                                                        & 39.50                                                                                        & 39.41                                                                    \\
\multicolumn{1}{l}{PointContrast~\cite{xie2020pointcontrast}}                                                                 & \multicolumn{1}{c}{ECCV 2020}                   & 21.90                                                                                           & 32.50                                                                                        & 41.10                                                                                        & -                                                                        \\
\multicolumn{1}{l}{DepthContrast~\cite{Zhang_2021_depthcontrast}}                                                                 & \multicolumn{1}{c}{ICCV 2021}                   & 22.10                                                                                           & 31.70                                                                                        & 41.50                                                                                        & -                                                                        \\
\multicolumn{1}{l}{PPKT~\cite{liu2021ppkt}}                                                                          & \multicolumn{1}{c}{arVix 2021}                  & 35.90                                                                                           & 37.52                                                                                        & 44.00                                                                                        & -                                                                        \\
\multicolumn{1}{l}{SLidR~\cite{sautier2022slidr}}                                                                         & \multicolumn{1}{c}{CVPR 2022}                   & 38.40                                                                                           & 38.83                                                                                        & 43.96                                                                                        & 47.12                                                                    \\
\multicolumn{1}{l}{ST-SLidR}                                                                      & \multicolumn{1}{c}{-}                          & \textbf{40.48}                                                                                           & \textbf{40.75}                                                                                        & \textbf{44.72}                                                                                        & \textbf{47.93}                                                                    \\ \midrule
\rowcolor{LightCyan}\multicolumn{2}{c}{\textit{Improvement}}                                                                                                                      & \textit{+2.08}                                                                                            & \textit{+1.92}                                                                                         & \textit{+0.76}                                                                                         & \textit{+0.81}        \\          \bottomrule                                                  
\end{tabular}}
\caption{Semantic segmentation results on nuScenes~\cite{caesar2020nuscenes}, SemanticKITTI~\cite{behley2019iccv_semkitti}  and Waymo~\cite{sun2020waymo} validation sets using 3D self-supervised methods. On nuScenes, we evaluate linear probing using 100$\%$ of the annotated training set and for finetuning, we evaluate using 1$\%$ of the data. In addition, we evaluate out-of-distribution performance on SemanticKITTI and Waymo datasets using 1$\%$ of the training set. All models are pretrained using the nuScenes dataset.} 
\label{tab:main_results}
\end{table}

\subsection{Evaluation}
To assess the quality of the pre-trained 3D representations, a point-wise linear classifier is added to the output of $f_{\theta_P}$. Two protocols are used to evaluate the performance on a semantic segmentation task, linear probing~\cite{alain2016understanding_linear_prob}, and fine-tuning. For linear probing, the parameters of $f_{\theta_P}$ are frozen and only the classifier head is trained on 100\% of the training data from the nuScenes dataset. Since gradients are not propagating back to $f_{\theta_P}$, the linear probe protocol directly evaluates the quality of the pre-trained representations. We evaluate the performance of the linear probing protocol on the nuScenes validation dataset.

For fine-tuning, the representations are evaluated under a low sample count, where the model is finetuned end-to-end using only 1\% of the annotated training data. The fine-tuning protocol allows us to study the utility of the pre-trained representations under a limited annotation budget. We study the finetuning performance on the nuScenes validation dataset and also evaluate the utility of the learned pre-trained nuScenes representations during finetuning on out-of-distribution data from the SemanticKITTI~\cite{behley2019iccv_semkitti} and Waymo~\cite{sun2020waymo} datasets. The number of classes for nuScenes, SemanticKITTI and Waymo datasets are 16, 19 and 22 respectively. The results are reported on the official validation sets of all datasets. For training, we use a linear combination of the Lovasz and cross-entropy loss, and the same training hyperparameters as SLidR.

\begin{table}[t]
\centering
\resizebox{\columnwidth}{!}{
\begin{tabular}{lcccc}
\toprule
\multicolumn{1}{c}{\multirow{3}{*}{\begin{tabular}[c]{@{}c@{}}3D\\ Initialization\end{tabular}}} & \multicolumn{1}{c}{\multirow{3}{*}{\begin{tabular}[c]{@{}c@{}}2D\\ Initialization\end{tabular}}} & \multicolumn{2}{c}{nuScenes}                                                                                                                                                                  & KITTI                                                                   \\ \cmidrule{3-5} 
\multicolumn{1}{c}{}                                                                             & \multicolumn{1}{c}{}                                                                             & \multicolumn{1}{c}{\multirow{2}{*}{\begin{tabular}[c]{@{}c@{}}Lin. Prob\\ 100\%\end{tabular}}} & \multicolumn{1}{c}{\multirow{2}{*}{\begin{tabular}[c]{@{}c@{}}Finetune\\ 1\%\end{tabular}}} & \multirow{2}{*}{\begin{tabular}[c]{@{}c@{}}Finetune\\ 1\%\end{tabular}} \\
\multicolumn{1}{c}{}                                                                             & \multicolumn{1}{c}{}                                                                             & \multicolumn{1}{c}{}                                                                           & \multicolumn{1}{c}{}                                                                        &                                                                         \\ \midrule
\multicolumn{1}{l}{SLidR}                                                                        & \multicolumn{1}{c}{\multirow{2}{*}{MoCoV2~\cite{chen2020_mocov2}}}                                                      & 38.40                                                                                           & 38.83                                                                                        & 43.96                                                                   \\
\multicolumn{1}{l}{ST-SLidR}                                                                     & \multicolumn{1}{c}{}                                                                             & \textbf{40.48}                                                                                           & \textbf{40.75}                                                                                        & \textbf{44.72}                                                                   \\ \midrule
\rowcolor{LightCyan}\multicolumn{2}{c}{\textit{Improvement}}                                                                                                                                                                       & \textit{+2.08}                                                                                            & \textit{+1.92}                                                                                         & \textit{+0.76}                                                                    \\ \midrule
\multicolumn{1}{l}{SLidR}                                                                        & \multicolumn{1}{c}{\multirow{2}{*}{SwAV~\cite{caron2020_swav}}}                                                        & 39.11                                                                                           & 38.81                                                                                        & 44.15                                                                   \\
\multicolumn{1}{l}{ST-SLidR}                                                                     & \multicolumn{1}{c}{}                                                                             & \textbf{40.49}                                                                                           & \textbf{40.86}                                                                                        & \textbf{44.98}                                                                   \\ \midrule
\rowcolor{LightCyan}\multicolumn{2}{c}{\textit{Improvement}}                                                                                                                                                                       & \textit{+1.38}                                                                                            & \textit{+2.05}                                                                                         & \textit{+0.83}                                                                    \\ \midrule
\multicolumn{1}{l}{SLidR}                                                                        & \multicolumn{1}{c}{\multirow{2}{*}{DINO~\cite{caron2021emerging_dino}}}                                                        & 37.76                                                                                           & 38.57                                                                                        & 43.94                                                                   \\
\multicolumn{1}{l}{ST-SLidR}                                                                     & \multicolumn{1}{c}{}                                                                             & \textbf{40.31}                                                                                           & \textbf{40.52}                                                                                        & \textbf{44.24}                                                                   \\ \midrule
\rowcolor{LightCyan}\multicolumn{2}{c}{\textit{Improvement}}                                                                                                                                                                       & \textit{+2.55}                                                                                            & \textit{+1.95}                                                                                         & \textit{+0.30}                                                                    \\ \midrule
\multicolumn{1}{l}{SLidR}                                                                        & \multicolumn{1}{c}{\multirow{2}{*}{OBoW~\cite{gidaris2021obow}}}                                                        & 36.56                                                                                           & 38.55                                                                                        & 43.84                                                                   \\
\multicolumn{1}{l}{ST-SLidR}                                                                     & \multicolumn{1}{c}{}                                                                             & \textbf{39.81}                                                                                           & \textbf{40.08}                                                                                        & \textbf{44.58}                                                                   \\ \midrule
\rowcolor{LightCyan}\multicolumn{2}{c}{\textit{Improvement}}                                                                                                                                                                       & \textit{+3.25}                                                                                            & \textit{+1.53}                                                                                         & \textit{+0.74}                                                                    \\ \midrule
\multicolumn{1}{l}{SLidR}                                                                        & \multicolumn{1}{c}{\multirow{2}{*}{DenseCL~\cite{wang2021_densecl}}}                                                     & 34.90                                                                                           & 37.50                                                                                        & \textbf{43.44}                                                                   \\
\multicolumn{1}{l}{ST-SLidR}                                                                     & \multicolumn{1}{c}{}                                                                             & \textbf{36.93}                                                                                           & \textbf{39.21}                                                                                        & 43.34                                                                   \\ \midrule
\rowcolor{LightCyan}\multicolumn{2}{c}{\textit{Improvement}}                                                                                                                                                                       & \textit{+2.03}                                                                                            & \textit{+1.71}                                                                                         & \textcolor{red}{\textit{-0.10}}       \\
\bottomrule
\end{tabular}}
\caption{The performance of vanilla SLidR compared to our proposed loss (ST-SlidR) when using different pretrained 2D self-supervised  models during training.} \label{tab:generalization_2d_pretrained_models}
\end{table}

\subsection{Results}
\noindent\textbf{Comparison with Baselines}. In~\Cref{tab:main_results}, we present the performance of Random initialization, PointContrast~\cite{xie2020pointcontrast}, and DepthContrast~\cite{Zhang_2021_depthcontrast} reported in ~\cite{sautier2022slidr}. For PPKT~\cite{liu2021ppkt}, SLidR~\cite{sautier2022slidr} and ST-SLidR, we run 3 pre-training experiments using SLidR's code base and report the mean performance for each setting. We observe that models initialized using weights from 3D SSL frameworks provide significant improvements over random initialization. In addition, in an outdoor setting where the density of the point cloud falls off rapidly at mid-to-long range, 3D SSL methods using superpixel-driven loss like SLidR~\cite{sautier2022slidr} leads to improved performance compared to point-level losses like PointContrast~\cite{xie2020pointcontrast} and PPKT~\cite{liu2021ppkt} or scene-level losses like DepthContrast~\cite{Zhang_2021_depthcontrast}. Pre-training using our proposed semantically-tolerant and balanced loss ST-SLidR provides a significant mIoU improvement of +2.08\% for linear probing and +1.92\% for few-shot fine-tuning tasks over state-of-the-art SLidR on nuScenes datasets. In addition, compared to SLidR, ST-SLidR also achieves better generalization in out-of-distribution few-shot semantic segmentation on SemanticKITTI and Waymo datasets. We also show that our proposed loss can improve the quality of 3D representations of pixel-to-point contrastive losses such as PPKT~\cite{liu2021ppkt} (see Table 7 in Appendix)

\vspace{0.05in}
\noindent\textbf{2D SSL Frameworks}. In~\Cref{tab:generalization_2d_pretrained_models}, we present results for experiments  using weights pre-trained with different 2D SSL frameworks.  We observe that ST-SLidR provides significant improvements over SLidR  of at least 1.38\% using linear probing and 1.5\% for fine-tuning on nuscenes dataset across all 2D pretrained models. This shows the robustness of ST-SLidR to selection of the 2D SSL pretrained model, indicating real benefit to feature transfer from 2D to 3D point encoders.
\begin{table}[t]
\centering
\resizebox{\columnwidth}{!}{
\begin{tabular}{lccccc}
\toprule
3D Initialization & 1\%   & 5\%   & 10\%  & 25\%  & 100\% \\ \midrule
Random            & 28.64 & 47.84 & 56.15 & 65.48 & 74.66 \\
SLidR             & 38.83 & 52.49 & 59.84 & 66.91 & 74.79 \\
ST-SLidR          & \textbf{40.75} & \textbf{54.69} & \textbf{60.75} & \textbf{67.70}  & \textbf{75.14} \\ \midrule
\rowcolor{LightCyan}\textit{Improvement}       & \textit{+1.92}  & \textit{+2.20}  & \textit{+0.91}  & \textit{+0.79}  & \textit{+0.35} \\
\bottomrule
\end{tabular}}
\caption{Finetune results on nuScenes as a function of the percentage of annotated data. Improvements are shown with respect to SLidR~\cite{sautier2022slidr}} \label{tab:annotation_efficiency}
\end{table}

\vspace{0.05in}
\noindent\textbf{Annotation Efficiency}. In~\Cref{tab:annotation_efficiency} we present results on the utility of the pre-trained representations as a function of the percentage of nuScenes training set. Here, we use the same training parameters selected by SLidR~\cite{sautier2022slidr} including learning rate and number of training epochs to evaluate the semantic segmentation fine-tuning performance of SLidR and ST-SLidR as a function of the number of labelled scenes. We observe that ST-SLidR outperforms SLidR by +1.92\%, +2.20\%, +0.91\%, +0.79\% and +0.35\% when fine-tuning on 1\%, 5\%, 10\%, 25\% and 100\% of the dataset, respectively.
\begin{table}[t]
\centering
\begin{tabular}{lcccc}
\toprule
\multirow{2}{*}{Method} & \multicolumn{2}{c}{Lin. Prob 100\%}                    & \multicolumn{2}{c}{Finetune 1\%}   \\ \cmidrule{2-5} 
                        & min & \multicolumn{1}{c}{maj} & min & maj \\ \midrule
SLidR                   & 27.46          & \textbf{62.47}                                  & 22.96          & 73.75             \\
ST-SLidR                & \textbf{30.64}          & 62.15                                  & \textbf{25.56}          & \textbf{74.15}             \\ \midrule
\rowcolor{LightCyan}\textit{Improvement}             & \textit{+3.18}           & \textcolor{red}{\textit{-0.32}}                                  & \textit{+2.61}           & \textit{+0.40}             \\
\bottomrule
\end{tabular}
\caption{We report mIOU for minority (min) and majority (maj) classes of nuScenes dataset. We group classes based on whether their superpixels occupy more than \textbf{5\%} of the superpixels in nuScenes training set.} \label{tab:class_based_IOU}
\end{table}

\begin{table}[t]
\centering
\begin{tabular}{cccc}
\toprule
\multicolumn{1}{c}{\multirow{2}{*}{\begin{tabular}[c]{@{}c@{}}Sematic Tolerant\\  Loss\end{tabular}}} & \multirow{2}{*}{\begin{tabular}[c]{@{}c@{}}Class Balanced\\  Loss\end{tabular}} & \multicolumn{1}{c}{\multirow{2}{*}{\begin{tabular}[c]{@{}c@{}}Lin. Prob\\ 100\%\end{tabular}}} & \multirow{2}{*}{\begin{tabular}[c]{@{}c@{}}Finetune\\ 1\%\end{tabular}} \\
\multicolumn{1}{c}{}                                                                               &                                                                                 & \multicolumn{1}{c}{}                                                                           &                                                                         \\ \midrule
\xmark                                                                                                  & \xmark                                                                              & 37.87                                                                                           & 38.96                                                                   \\
\xmark                                                                                                  & \cmark                                                                             & 38.33                                                                                           & 39.73                                                                   \\
\cmark                                                                                                 & \xmark                                                                              & 40.04                                                                                           & 40.19                                                                   \\
\cmark                                                                                                 & \cmark                                                                             & \textbf{40.48}                                                                                           & \textbf{40.75}                \\
\toprule
\end{tabular}
\caption{Contribution of semantic awareness and class agnostic balancing on ST-SLidR.} \label{tab:loss_components}
\end{table}

\begin{table}[t]
\centering
\begin{tabular}{lcc}
\toprule
\multicolumn{1}{c}{\multirow{2}{*}{Loss}} & \multicolumn{1}{c}{\multirow{2}{*}{\begin{tabular}[c]{@{}c@{}}Lin. Prob 100\% \\ MoCoV2\end{tabular}}} & \multirow{2}{*}{\begin{tabular}[c]{@{}c@{}}Lin. Prob 100\% \\ SwAV\end{tabular}} \\
\multicolumn{1}{c}{}                      & \multicolumn{1}{c}{}                                                                                      &                                                                                   \\ \midrule
$\alpha_{min} = 0.0$                                & 37.99                                                                                                      &   36.21                                                                           \\
$\alpha_{min} = 0.2$                                & 38.64                                                                                                      &   36.86                                                                           \\
$\alpha_{min} = 0.5$                                & 39.48                                                                                                      &  40.03                                                                            \\
$\alpha_{min} = 0.8$                                & 38.23                                                                                                      & 39.15                                                                             \\ \midrule
$K=1\%$                                      & 40.04                                                                                                      &  40.42                                                                          \\
$K=5\%$                                      & \textbf{40.38}                                                                                              & \textbf{40.84}                                                                                    \\
$K=10\%$                                     & 40.35                                                                                                      &  40.05 \\                                                                            
\bottomrule
\end{tabular}
\caption{Comparison between similarity-aware $\mathcal{L}_\alpha$ versus nearest-neighbour-aware $\mathcal{L}_{knn}$ loss. Here, we report mIOU on the validation set of nuScenes~\cite{caesar2020nuscenes} set.} \label{tab:cosine_sim_K_nearest}
\vspace{-5mm}
\end{table}

\vspace{0.05in}
\noindent\textbf{Class Imbalance}. We conduct an experiment to study which semantic classes gain the most from the semantically tolerant contrastive loss. We compute the percentage of superpixels for each semantic class in the nuScenes pre-training set. Then, we create two sets of classes. The minority set contains all classes with fewer than 5\% of the superpixels in the pre-training set. The remaining classes are added to the majority set. Out of the 16 classes in the nuScenes dataset, 11 classes are categorized as minority classes. In~\Cref{tab:class_based_IOU}, the mean IoU for minority and majority sets for linear probing and fine-tuning on nuScenes validation set is presented. Compared to SLidR, ST-SLidR learns representations that significantly improve segmentation performance by +3.18\% for linear probing and +2.61\% for fine-tuning on the 11 minority classes. Interestingly, the significant improvement on minority classes comes with almost no degradation on majority classes.

\subsection{Ablations}
\noindent\textbf{Contribution of Loss Components}. We conduct ablation studies to validate the contribution of the two components of ST-SLidR. Here, semantic tolerant loss denotes $\mathcal{L}_{knn}$ with K set to 1\% of the mini-batch. ~\Cref{tab:loss_components} shows that (1) Both semantic tolerant and class balancing can improve the quality of the learned representation on their own, (2) Semantic tolerant loss significantly improves the linear separability of the 3D representations as fewer false negative samples contribute to the loss, (3) Using both components achieves the best performance on linear probing and few-shot semantic segmentation.  

\vspace{0.05in}
\noindent\textbf{Similarity versus Nearest-Neighbour-aware Loss}. We conduct an ablation study to show the quality of the 3D representations for similarity-aware loss $\mathcal{L}_{\alpha}$ and nearest neighbour-aware loss $\mathcal{L}_{knn}$. For pre-training experiments using $\mathcal{L}_{\alpha}$, we vary the minimum similarity threshold $\alpha_{min}$ and for $\mathcal{L}_{knn}$, we vary percentage of top-$K$ nearest neighbours to be excluded from the set of negative samples.~\Cref{tab:cosine_sim_K_nearest} shows pre-training with $\mathcal{L}_{knn}$ results in 3D representations that are much more linearly separable than $\mathcal{L}_{\alpha}$.  

\section{Conclusion}
\label{sec:conclusion}
We present a novel 2D-to-3D representation learning framework for autonomous driving datasets that reduces the contribution of false negative samples by explicitly considering the similarity of self-supervised image features. In addition we propose balancing the pretraining between over and under-represented samples by using aggregate sample-to-samples similarity as a proxy for class imbalance. Our proposed contributions are shown to additively improve common 2D-to-3D representation learning methods in all evaluation settings on 3D semantic segmentation, especially for under-represented classes.
{\small
\bibliographystyle{ieee_fullname}
\bibliography{egbib}

\begin{thebibliography}{10}\itemsep=-1pt

\bibitem{slic}
Radhakrishna Achanta, Appu Shaji, Kevin Smith, Aurelien Lucchi, Pascal Fua, and
  Sabine Süsstrunk.
\newblock Slic superpixels compared to state-of-the-art superpixel methods.
\newblock {\em IEEE TPAMI}, 34(11):2274--2282, 2012.

\bibitem{alain2016understanding_linear_prob}
Guillaume Alain and Yoshua Bengio.
\newblock Understanding intermediate layers using linear classifier probes.
\newblock {\em arXiv preprint arXiv:1610.01644}, 2016.

\bibitem{bachman2019learning}
Philip Bachman, R~Devon Hjelm, and William Buchwalter.
\newblock Learning representations by maximizing mutual information across
  views.
\newblock {\em NeurIPS}, 32, 2019.

\bibitem{bardes2022vicreg}
Adrien Bardes, Jean Ponce, and Yann Lecun.
\newblock Vicreg: Variance-invariance-covariance regularization for
  self-supervised learning.
\newblock In {\em ICLR}, 2022.

\bibitem{behley2019iccv_semkitti}
J. Behley, M. Garbade, A. Milioto, J. Quenzel, S. Behnke, C. Stachniss, and J.
  Gall.
\newblock {SemanticKITTI: A Dataset for Semantic Scene Understanding of LiDAR
  Sequences}.
\newblock In {\em ICCV}, 2019.

\bibitem{caesar2020nuscenes}
Holger Caesar, Varun Bankiti, Alex~H Lang, Sourabh Vora, Venice~Erin Liong,
  Qiang Xu, Anush Krishnan, Yu Pan, Giancarlo Baldan, and Oscar Beijbom.
\newblock nuscenes: A multimodal dataset for autonomous driving.
\newblock In {\em CVPR}, pages 11621--11631, 2020.

\bibitem{caron2020_swav}
Mathilde Caron, Ishan Misra, Julien Mairal, Priya Goyal, Piotr Bojanowski, and
  Armand Joulin.
\newblock Unsupervised learning of visual features by contrasting cluster
  assignments.
\newblock {\em NeurIPS}, 33:9912--9924, 2020.

\bibitem{caron2021emerging_dino}
Mathilde Caron, Hugo Touvron, Ishan Misra, Herv{\'e} J{\'e}gou, Julien Mairal,
  Piotr Bojanowski, and Armand Joulin.
\newblock Emerging properties in self-supervised vision transformers.
\newblock In {\em ICCV}, pages 9650--9660, 2021.

\bibitem{chen2020_simclr}
Ting Chen, Simon Kornblith, Mohammad Norouzi, and Geoffrey Hinton.
\newblock A simple framework for contrastive learning of visual
  representations.
\newblock In {\em ICML}, pages 1597--1607, 2020.

\bibitem{chen2020_mocov2}
Xinlei Chen, Haoqi Fan, Ross Girshick, and Kaiming He.
\newblock Improved baselines with momentum contrastive learning.
\newblock {\em arXiv preprint arXiv:2003.04297}, 2020.

\bibitem{gidaris2021obow}
Spyros Gidaris, Andrei Bursuc, Gilles Puy, Nikos Komodakis, Matthieu Cord, and
  Patrick Perez.
\newblock Obow: Online bag-of-visual-words generation for self-supervised
  learning.
\newblock In {\em CVPR}, pages 6830--6840, 2021.

\bibitem{he2016deep_resnet}
Kaiming He, Xiangyu Zhang, Shaoqing Ren, and Jian Sun.
\newblock Deep residual learning for image recognition.
\newblock In {\em CVPR}, pages 770--778, 2016.

\bibitem{henaff2021efficient_detcon}
Olivier~J H{\'e}naff, Skanda Koppula, Jean-Baptiste Alayrac, Aaron Van~den
  Oord, Oriol Vinyals, and Jo{\~a}o Carreira.
\newblock Efficient visual pretraining with contrastive detection.
\newblock In {\em ICCV}, pages 10086--10096, 2021.

\bibitem{Hu_2022_PDV}
Jordan S.~K. Hu, Tianshu Kuai, and Steven~L. Waslander.
\newblock Point density-aware voxels for lidar 3d object detection.
\newblock In {\em CVPR}, pages 8469--8478, June 2022.

\bibitem{jiang2021guided}
Li Jiang, Shaoshuai Shi, Zhuotao Tian, Xin Lai, Shu Liu, Chi-Wing Fu, and Jiaya
  Jia.
\newblock Guided point contrastive learning for semi-supervised point cloud
  semantic segmentation.
\newblock In {\em ICCV}, pages 6423--6432, 2021.

\bibitem{li2022closer}
Lanxiao Li and Michael Heizmann.
\newblock A closer look at invariances in self-supervised pre-training for 3d
  vision.
\newblock {\em ECCV}, 2022.

\bibitem{liu2021ppkt}
Yueh-Cheng Liu, Yu-Kai Huang, Hung-Yueh Chiang, Hung-Ting Su, Zhe-Yu Liu,
  Chin-Tang Chen, Ching-Yu Tseng, and Winston~H Hsu.
\newblock Learning from 2d: Contrastive pixel-to-point knowledge transfer for
  3d pretraining.
\newblock {\em arXiv preprint arXiv:2104.04687}, 2021.

\bibitem{DVF}
Anas Mahmoud, Jordan S.~K. Hu, and Steven~L. Waslander.
\newblock Dense voxel fusion for 3d object detection.
\newblock In {\em Proceedings of the IEEE/CVF Winter Conference on Applications
  of Computer Vision (WACV)}, pages 663--672, January 2023.

\bibitem{bounding_box_painting}
Anas Mahmoud and Steven~L. Waslander.
\newblock Sequential fusion via bounding box and motion pointpainting for 3d
  objection detection.
\newblock In {\em CRV}, 2021.

\bibitem{russakovsky2015imagenet}
Olga Russakovsky, Jia Deng, Hao Su, Jonathan Krause, Sanjeev Satheesh, Sean Ma,
  Zhiheng Huang, Andrej Karpathy, Aditya Khosla, Michael Bernstein, et~al.
\newblock Imagenet large scale visual recognition challenge.
\newblock {\em IJCV}, 115(3):211--252, 2015.

\bibitem{sautier2022slidr}
Corentin Sautier, Gilles Puy, Spyros Gidaris, Alexandre Boulch, Andrei Bursuc,
  and Renaud Marlet.
\newblock Image-to-lidar self-supervised distillation for autonomous driving
  data.
\newblock In {\em CVPR}, pages 9891--9901, 2022.

\bibitem{sun2020waymo}
Pei Sun, Henrik Kretzschmar, Xerxes Dotiwalla, Aurelien Chouard, Vijaysai
  Patnaik, Paul Tsui, James Guo, Yin Zhou, Yuning Chai, Benjamin Caine, et~al.
\newblock Scalability in perception for autonomous driving: Waymo open dataset.
\newblock In {\em CVPR}, pages 2446--2454, 2020.

\bibitem{tsne}
Laurens Van~der Maaten and Geoffrey Hinton.
\newblock Visualizing data using t-sne.
\newblock {\em Journal of machine learning research}, 9(11), 2008.

\bibitem{wang2021understanding}
Feng Wang and Huaping Liu.
\newblock Understanding the behaviour of contrastive loss.
\newblock In {\em CVPR}, pages 2495--2504, 2021.

\bibitem{wang2021_densecl}
Xinlong Wang, Rufeng Zhang, Chunhua Shen, Tao Kong, and Lei Li.
\newblock Dense contrastive learning for self-supervised visual pre-training.
\newblock In {\em CVPR}, pages 3024--3033, 2021.

\bibitem{wu2018unsupervised_instance}
Zhirong Wu, Yuanjun Xiong, Stella~X Yu, and Dahua Lin.
\newblock Unsupervised feature learning via non-parametric instance
  discrimination.
\newblock In {\em CVPR}, pages 3733--3742, 2018.

\bibitem{xie2020pointcontrast}
Saining Xie, Jiatao Gu, Demi Guo, Charles~R Qi, Leonidas Guibas, and Or Litany.
\newblock Pointcontrast: Unsupervised pre-training for 3d point cloud
  understanding.
\newblock In {\em ECCV}, pages 574--591. Springer, 2020.

\bibitem{zbontar2021barlow}
Jure Zbontar, Li Jing, Ishan Misra, Yann LeCun, and St{\'e}phane Deny.
\newblock Barlow twins: Self-supervised learning via redundancy reduction.
\newblock In {\em ICML}, pages 12310--12320. PMLR, 2021.

\bibitem{Zhang_2021_depthcontrast}
Zaiwei Zhang, Rohit Girdhar, Armand Joulin, and Ishan Misra.
\newblock Self-supervised pretraining of 3d features on any point-cloud.
\newblock In {\em ICCV}, pages 10252--10263, October 2021.

\bibitem{zhu2021cylindrical}
Xinge Zhu, Hui Zhou, Tai Wang, Fangzhou Hong, Yuexin Ma, Wei Li, Hongsheng Li,
  and Dahua Lin.
\newblock Cylindrical and asymmetrical 3d convolution networks for lidar
  segmentation.
\newblock In {\em CVPR}, pages 9939--9948, 2021.

\end{thebibliography}
}
\clearpage
\appendix
\newpage
This supplementary material contains the following four sections. In Section~\ref{sec:PPKT}, we show that point-to-pixel level contrastive losses like PPKT~\cite{liu2021ppkt} can also benefit from the proposed semantically tolerant loss. In Section~\ref{sec:class_distribution},  we visualize the class imbalance in nuScenes dataset on the superpixel level and then report the per-class fine-tuning performance of SLidR and ST-SLidR represenations on nuscenes dataset. In Section~\ref{sec:similarity}, we visualize the superpixel-to-superpixel similarity across a range of 2D self-supervised pretrained models.  Finally, In Section~\ref{sec:limitations}, we discuss the  limitations of ST-SLidR.

\section{Semantically Tolerant PPKT}
\label{sec:PPKT}
We conduct an experiment to evaluate whether pixel-to-pixel semantic similarity can be used to improve the quality of learned representations of pixel-to-point contrastive losses like PPKT~\cite{liu2021ppkt}. The main challenge of utilizing  pixel-to-pixel similarity to infer false negative pixels, is the high level of noise compared to superpixel-to-superpixel similarity. 
Starting with the implementation of PPKT provided by SLidR's code base~\cite{sautier2022slidr}, we implement $\mathcal{L}_{knn}$ on the pixel level. We run two sets of experiments using 4096 and 8192 point-pixel pairs per batch. We use a batch size of 16. Here, we report the average of 3 runs for each experiment. In~\Cref{tab:PPKT}, we observe that semantically tolerant PPKT loss provides a modest improvement over PPKT~\cite{liu2021ppkt}. We also experimented with balancing PPKT using aggregate pixel-to-pixels similarity but due to the high level of noise in pixel level similarity, we did not observe any significant improvement. 
\section{Class Imbalance}
\label{sec:class_distribution}
\subsection{Superpixel Class Imbalance}
\begin{figure}[h]
  \centering
  \includegraphics[scale=0.3]{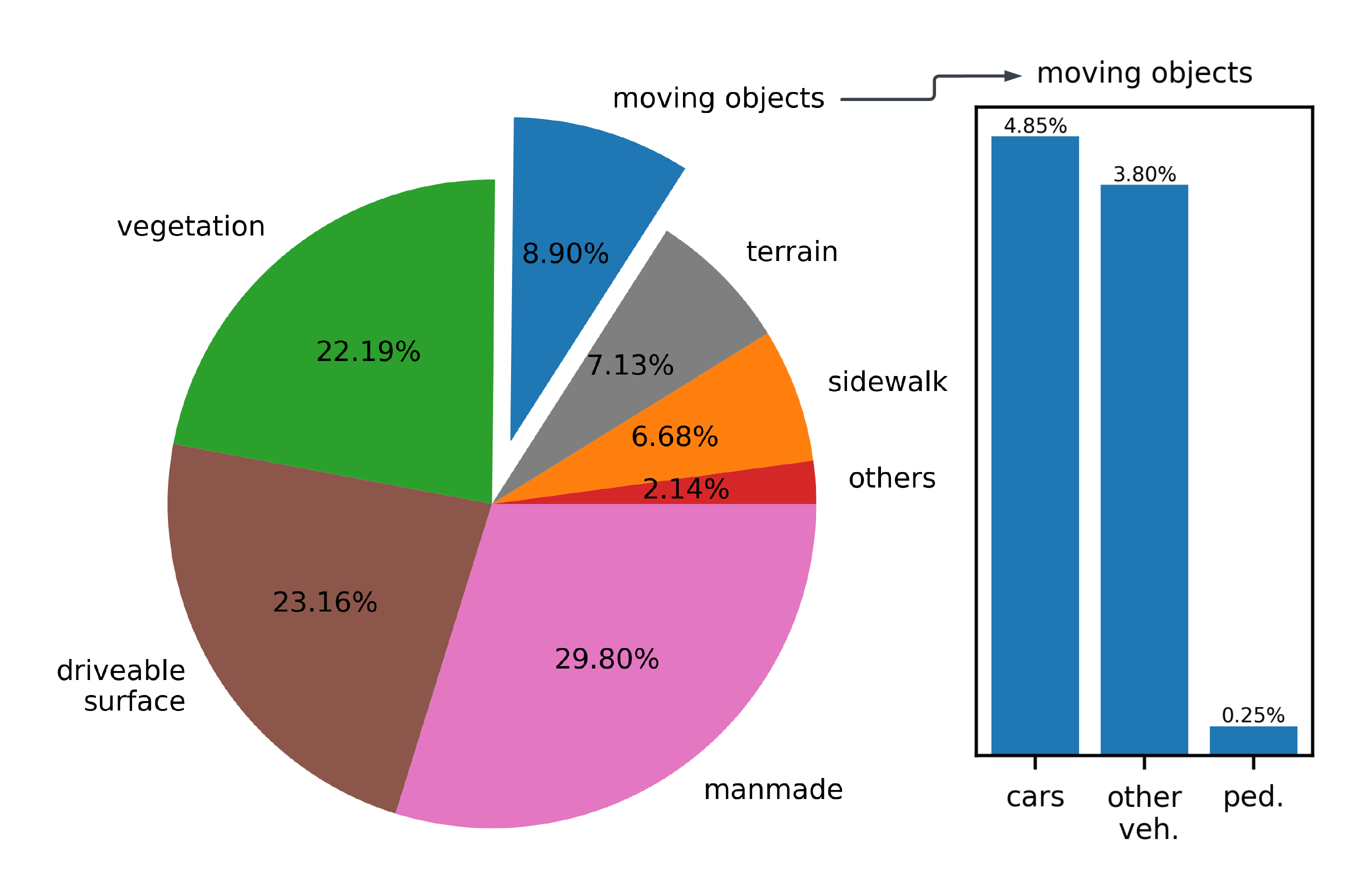}
  \caption{Class distribution of nuScenes dataset at superpixel level.}
  \label{fig:superpixel_class_imbalance}
\end{figure}

In~\Cref{fig:superpixel_class_imbalance}, we show the distribution of classes in the nuScenes~\cite{caesar2020nuscenes} dataset at the superpixel level. To determine the class of a superpixel, we first project the LiDAR point cloud onto the 2D image. The class of a superpixel is given by ground truth LiDAR point-wise labels of the points within the superpixel of interest. Specifically, its class is the same as its LiDAR points' label. In the cases where LiDAR points of multiple classes occur within a superpixel, we assign the class of the superpixel to be the mode of the points' LiDAR labels. We exclude the superpixels without LiDAR points as they are not used in pretraining. Note that the "others" category on the pie chart includes movable objects such as traffic cones and barriers. We observe that only 8.9\% of the superpixels cover moving objects like vehicles and pedestrians, while a large portion of the superpixels correspond to static classes like driveable surface, vegetation, and manmade. Thus, the pretraining loss of PPKT and SLidR is dominated by gradients from over-represented classes. It is important to note that accurately segmenting moving objects is critical for autonomous driving agents as they share the same environment and their actions will affect the agent. ST-SLidR specifically improves the quality of representations of minority classes which includes moving objects (see~\Cref{tab:class_based_IOU} and ~\Cref{tab:iou_per_class}). 
\begin{table}[t]
\centering
\begin{tabular}{lccc}
\toprule
\multicolumn{1}{l}{\multirow{3}{*}{Method}} & \multicolumn{1}{c}{\multirow{3}{*}{\begin{tabular}[c]{@{}c@{}}Number of \\ Samples\end{tabular}}} & \multicolumn{2}{c}{nuScenes}                                                                                                                                                                     \\ \cline{3-4} 
\multicolumn{1}{l}{}                        & \multicolumn{1}{c}{}                                                                              & \multicolumn{1}{l}{\multirow{2}{*}{\begin{tabular}[c]{@{}l@{}}Lin. Prob \\ 100\%\end{tabular}}} & \multicolumn{1}{l}{\multirow{2}{*}{\begin{tabular}[c]{@{}l@{}}Finetune  \\ 1\%\end{tabular}}} \\
\multicolumn{1}{l}{}                        & \multicolumn{1}{c}{}                                                                              & \multicolumn{1}{l}{}                                                                            & \multicolumn{1}{l}{}                                                                          \\ \hline
\multicolumn{1}{l}{PPKT}                    & \multicolumn{1}{c}{\multirow{2}{*}{4096}}                                                         & 35.90                                                                                            & 37.52                                                                                         \\
\multicolumn{1}{l}{ST-PPKT}                 & \multicolumn{1}{c}{}                                                                              & \textbf{36.70}                                                                                            & \textbf{38.32}                                                                                         \\ \midrule
\rowcolor{LightCyan}\multicolumn{2}{c}{\textit{Improvement}}                                                                                                                   & \textit{+0.80}                                                                                             & \textit{+0.80}                                                                                          \\ \midrule
\multicolumn{1}{l}{PPKT}                    & \multicolumn{1}{c}{\multirow{2}{*}{8192}}                                                         & 35.57                                                                                            & 38.01                                                                                         \\
\multicolumn{1}{l}{ST-PPKT}                 & \multicolumn{1}{c}{}                                                                              & \textbf{36.64}                                                                                            & \textbf{38.60}                                                                                         \\ \midrule
\rowcolor{LightCyan}\multicolumn{2}{c}{\textit{Improvement}}                                                                                                                   & \textit{+1.07}                                                                                             & \textit{+0.59}        \\                                                                                 
\bottomrule
\end{tabular}
\caption{Pixel-to-Pixel feature similarity used to remove the closest $k$ nearest negative pixels identified as false negatives. Here, we show results for 4096 and 8192 pixel-point contrastive pairs per batch. We report semantic segmentation results on nuScenes.} \label{tab:PPKT}
\end{table}
\begin{table*}[t]
\centering
\centering
\resizebox{\textwidth}{!}{
\begin{tabular}{lccccccccccccccccc}
\toprule
Method      & \rotatebox{90}{barrier} & \rotatebox{90}{bicycle} & \rotatebox{90}{bus}  & \rotatebox{90}{car}  & \rotatebox{90}{const. veh.} & \rotatebox{90}{motorcycle} & \rotatebox{90}{pedestrian} & \rotatebox{90}{traffic cone} & \rotatebox{90}{trailer} & \rotatebox{90}{truck} & \rotatebox{90}{driv. surf.} & \rotatebox{90}{other flat} & \rotatebox{90}{sidewalk} & \rotatebox{90}{terrain} & \rotatebox{90}{manmade} & \rotatebox{90}{vegetation} & mIoU \\ \hline
Random      & 0.0     & 0.0     & 8.1  & 65.0 & 0.1         & 6.6        & 21.0       & 9.0          & 9.3     & 25.8  & 89.5        & 14.8       & 41.7     & 48.7    & 72.4    & 73.3       & 30.3 \\
SLidR       & 0.0     & 1.8     & 15.4 & 73.1 & 1.9         & 19.9       & 47.2       & 17.1         & 14.5    & 34.5  & 92.0        & 27.1       & 53.6     & 61.0    & 79.8    & 82.3       & 38.8 \\
ST-SLidR    & 0.0     & \textbf{2.7}     & \textbf{16.0} & \textbf{74.5} & \textbf{3.2}         & \textbf{25.4}       & \textbf{50.9}       & \textbf{20.0}         & \textbf{17.7}   & \textbf{40.2}  & 92.0        & \textbf{30.7}       & \textbf{54.2}     & 61.1    & \textbf{80.5}    & \textbf{82.9}       & \textbf{40.7} \\ \hline
\rowcolor{LightCyan}\textit{Improvement} & \textit{+0.0}     & \textit{+0.8}     & \textit{+0.6}  & \textit{+1.4}  & \textit{+1.3}         & \textit{+5.5}        & \textit{+3.7}        & \textit{+2.9}          & \textit{+3.1}     & \textit{+5.7}   & \textit{+0.0}         & \textit{+3.6}        & \textit{+0.6}      & \textit{+0.0}     & \textit{+0.7}     & \textit{+0.6}        & \textit{+1.9} \\
\bottomrule
\end{tabular}}
\caption{Per-class 3D semantic segmentation using 1\% of labelled data for fine-tuning on nuscenes dataset on official validation set. We report the mean performance of 3 pretrained SLidR and ST-SLidR models.} \label{tab:iou_per_class}
\end{table*}
\begin{figure*}[t]
  \centering
  \includegraphics[scale=1.2]{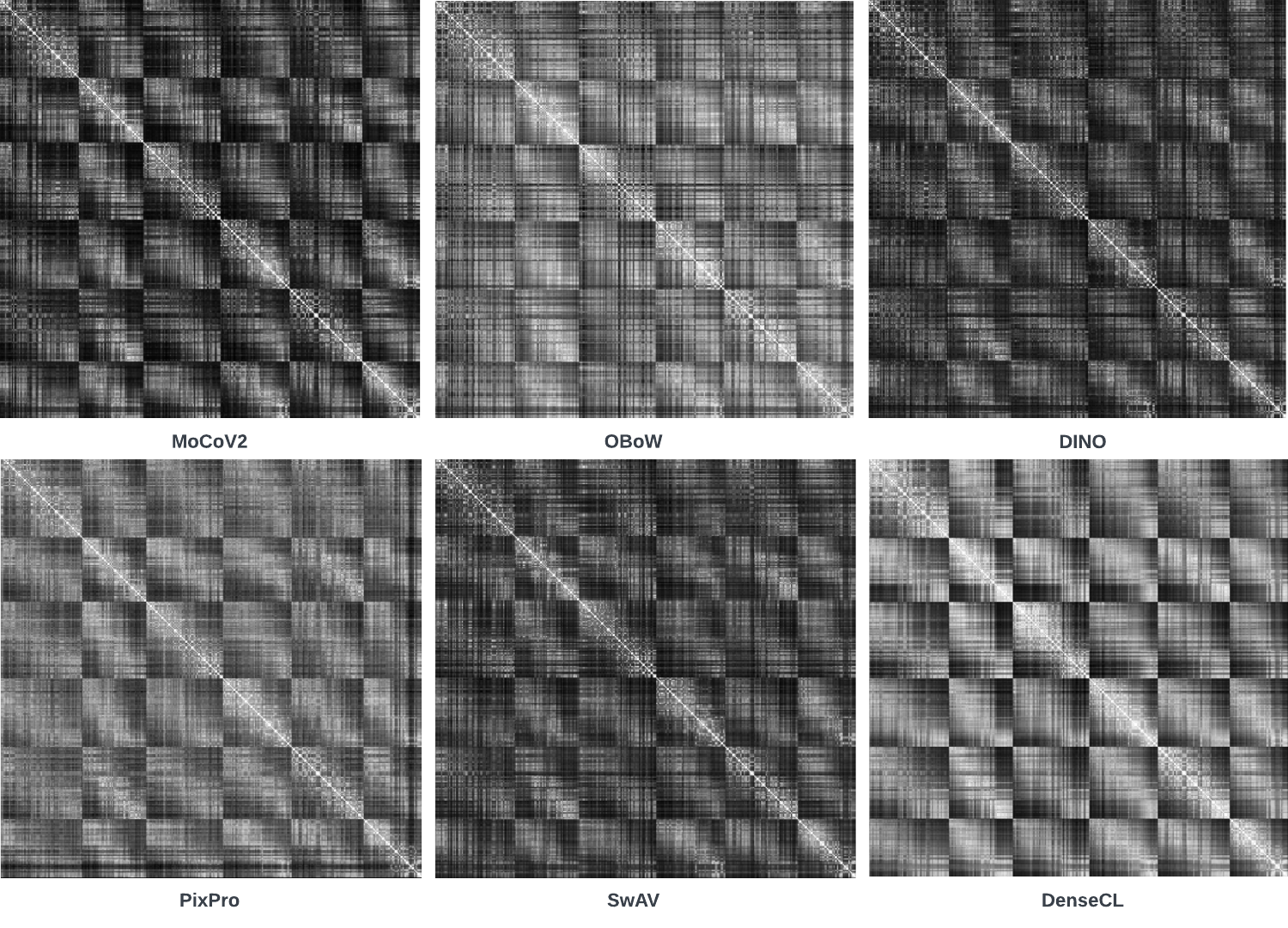}

  \caption{Superpixel-to-superpixel cosine similarity for an entire scene consisting of 6 cameras. Here, we show the similarity estimated using different self-supervised 2D pretrained models.}
  \label{fig:superpixel-similarity}
\end{figure*}

\begin{figure*}[t]
\centering
\begin{tabular}{cc}
\includegraphics[width=0.48\textwidth]{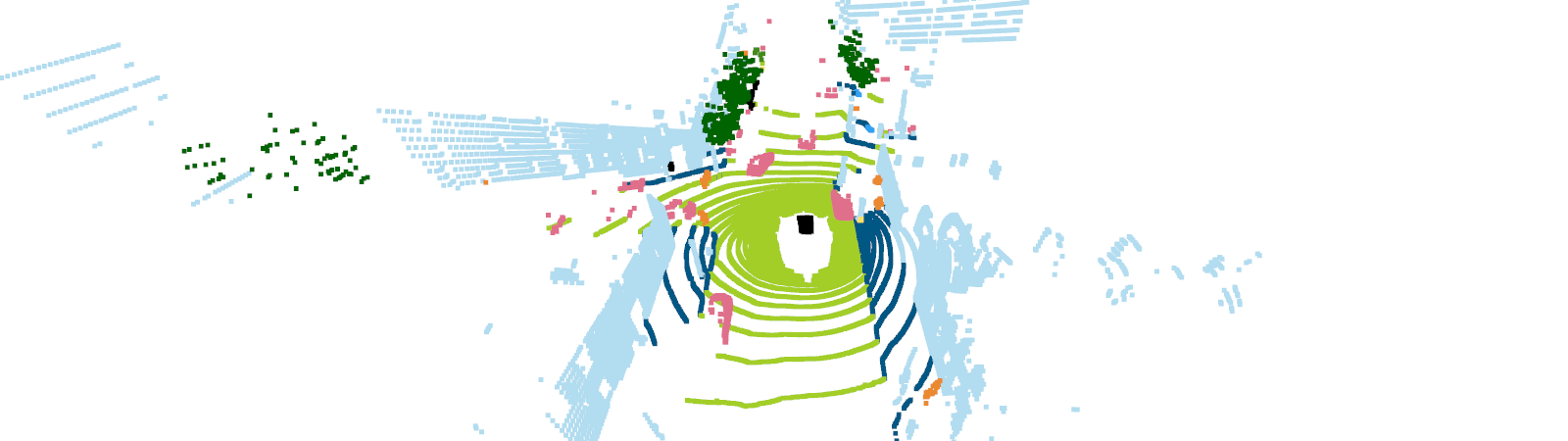} &
\includegraphics[width=0.48\textwidth]{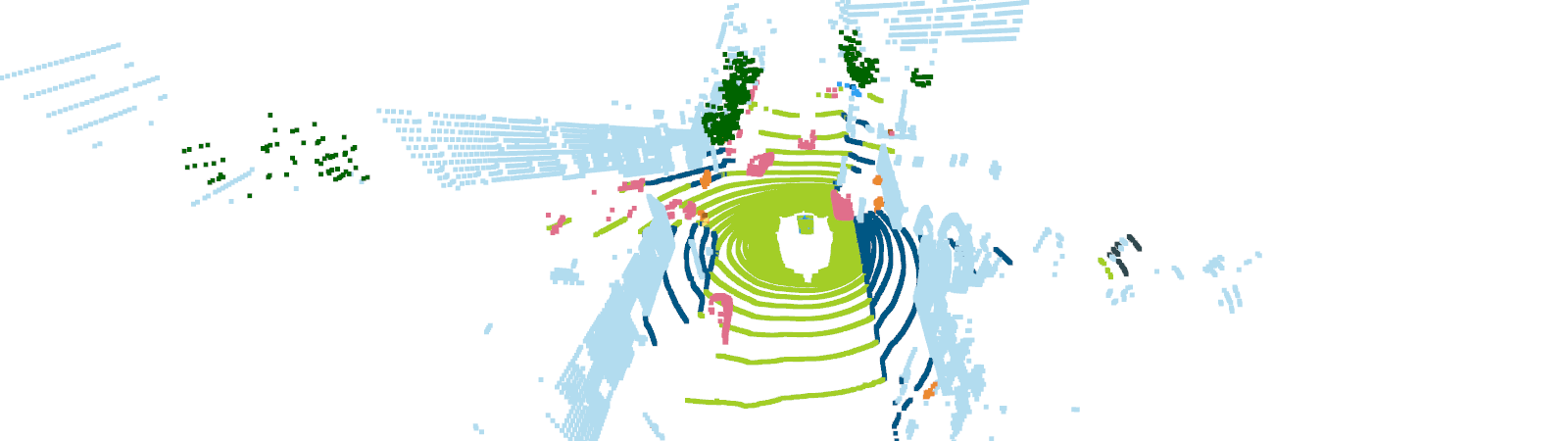} \\[8pt]
\includegraphics[width=0.48\textwidth]{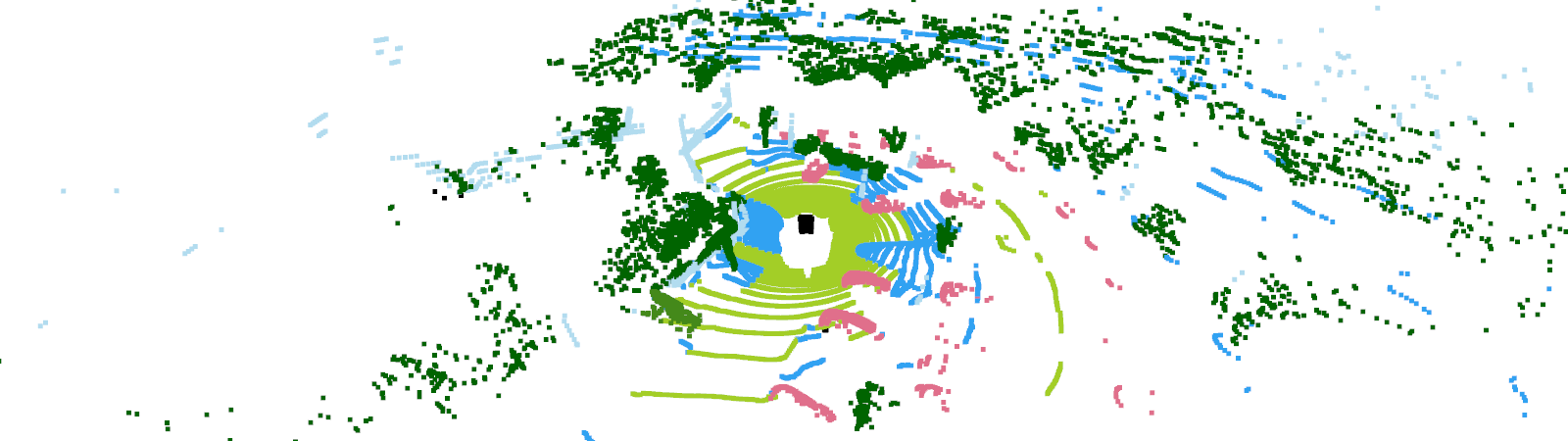} &
\includegraphics[width=0.48\textwidth]{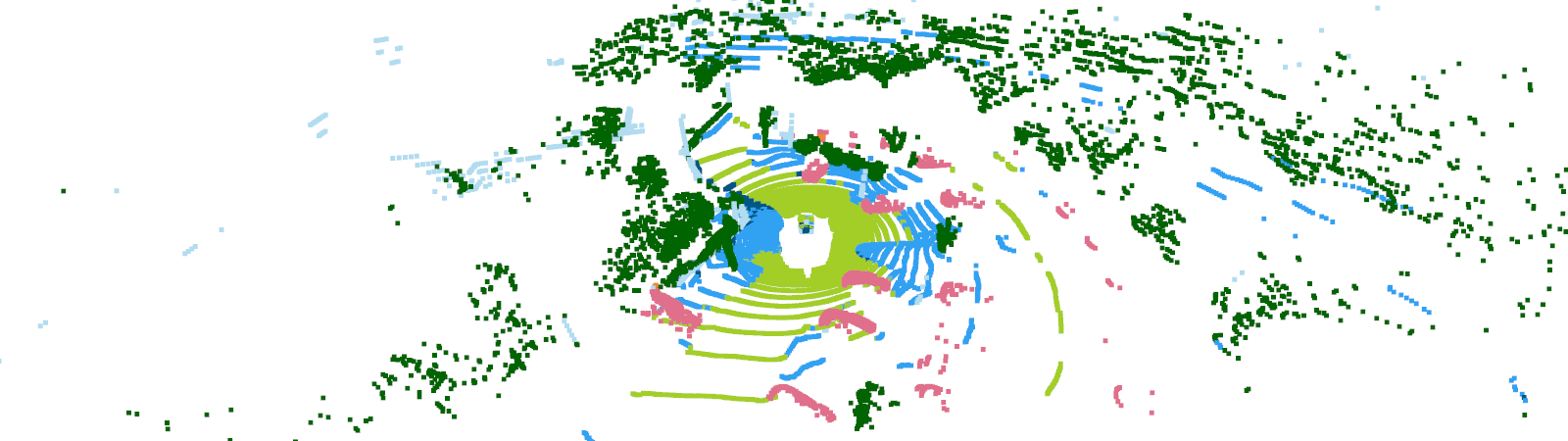}
\end{tabular}
  \caption{Ground truth (left) and ST-SLidR (right) segmentation results on the nuScenes dataset. ST-SLidR is finetuned on 1\% of the data.}
  \label{fig:segmentation-examples}
\end{figure*}
\subsection{Per-class Performance}
~\Cref{tab:iou_per_class} shows the average per-class performance fine-tuning SLidR and ST-SLidR on 1\% of nuscenes dataset. We observe that 3D representations learned by ST-SLidR significantly improve performance on minority classes like moving objects. For instance, we see an improvement of +5.5\% IoU on motorcyclists which consists of less than 0.04\% of the superpixels, +3.7\% IoU on pedestrians which consists of 0.25\% of the superpixels and 5.7\% IoU on trucks which consists of 2.11\% of the superpixels. 
\section{Superpixel Semantic Similarity}
\label{sec:similarity}
In~\Cref{fig:superpixel-similarity}, we use pretrained weights from multiple 2D SSL frameworks to extract superpixel features from the 6 cameras covering a single scene from nuscenes dataset. Then, we compute the superpixel-to-superpixel cosine similarity ranging from 0.0 (black) to 1.0 (white).~\Cref{fig:superpixel-similarity} shows that 2D SSL frameworks learn different representations, however, we can see that similarity patterns appear to be consistent across different frameworks. ST-SLidR assumes that the value of cosine similarity can be different across different pretrained models, but the relative of order of similarity with respect to an anchor is more consistent. This is demonstrated in~\Cref{tab:generalization_2d_pretrained_models}, where ST-SLidR provides significant gain over SLidR across multiple 2D pretrained models.
\section{Limitations}
\label{sec:limitations}
\subsection{Fixed Number of Negative Samples}
We address the issue of contrasting semantically similar point and image regions by excluding a subset of the closest negative samples to the anchor from the pool of negative samples. Since the K nearest neighbours are excluded, a fixed number of false negative samples are identified for each anchor. However,~\Cref{fig:superpixel_features} shows that the number of semantically similar samples greatly vary based on the semantic class of the anchor. For instance, the number of samples similar to a road or a vegetation anchor is much larger than the number of samples similar to a car or a pedestrian anchor. This is mainly due to the severe class imbalance in autonomous driving datasets. A potential solution for future work is to design an adaptive K nearest neigbour loss, where the value of K is a function of the aggregate sample-to-samples similarity. Over-represented anchors are similar to many negative samples in a batch and therefore the value of K should be higher for these anchors than under-represented anchors. 
\subsection{Frozen Image Encoder}
Authors in SLidR~\cite{sautier2022slidr} observe that backpropagating gradients to the image encoder can result in degenerate solutions, where the contrastive loss is easily minimized without learning useful 3D representations for downstream tasks. One of the advantages of updating the image encoder parameters initialized by ImageNet pretrained weights, is to learn optimal 2D features for autonomous driving scenes. To prevent degenerate solutions, the image encoder can be first initialized with ImageNet pretrained weights, and any 2D SSL framework can be used to learn optimal 2D representations for autonomous driving images. Finally, the image encoder is frozen and then ST-SLidR can be used to transfer knowledge from the 2D features to the point cloud encoder.

\end{document}